\documentclass[conference]{IEEEtran}
\IEEEoverridecommandlockouts
\usepackage{amsmath,amssymb,amsfonts}
\usepackage{algorithmic}
\usepackage{graphicx}
\usepackage{float}
\usepackage{textcomp}
\usepackage{biblatex}
\usepackage{xcolor}
\usepackage{multicol}
\usepackage{subcaption}
\usepackage{longtable}
\usepackage{lipsum}
\usepackage{booktabs, capt-of, tabularx}
\usepackage{cuted, stfloats}
\usepackage{hyperref}
\hypersetup{
    colorlinks=true,
    linkcolor=black,
    filecolor=black,
    citecolor=black,
    urlcolor=black,
}

\makeatletter
\renewcommand\subsubsection{\@startsection{subsubsection}{3}{\z@}%
  {-3.25ex\@plus -1ex \@minus -.2ex}%
  {1.5ex \@plus .2ex}%
  {\normalfont\normalsize}}
\makeatother

\begin{document}

\title{Universal Cross-Lingual Text Classification
}

\makeatletter
\newcommand{\newlineauthors}{%
  \end{@IEEEauthorhalign}\hfill\mbox{}\par
  \mbox{}\hfill\begin{@IEEEauthorhalign}
}
\makeatother

\author{
    \IEEEauthorblockN{Riya Savant\IEEEauthorrefmark{1}\IEEEauthorrefmark{2}, Anushka Shelke\IEEEauthorrefmark{1}\IEEEauthorrefmark{2}, Sakshi Todmal\IEEEauthorrefmark{1}\IEEEauthorrefmark{2}, Sanskruti Kanphade\IEEEauthorrefmark{1}\IEEEauthorrefmark{2}, Ananya Joshi\IEEEauthorrefmark{1}\IEEEauthorrefmark{2}, Raviraj Joshi\IEEEauthorrefmark{2}\IEEEauthorrefmark{3}}
    
    \IEEEauthorblockA{
        \IEEEauthorrefmark{1}\textit{MKSSS Cummins College of Engineering for Women, Pune}\\
        \IEEEauthorrefmark{2}\textit{L3Cube Labs, Pune}\\
        \IEEEauthorrefmark{3}\textit{Indian Institute of Technology, Madras}\\
        %Pune, India \\
        \{riya.savant, anushkashelke2003, sakshitodmal401, sanskruti.kn, joshiananya20\}@gmail.com, ravirajoshi@gmail.com
    }
}

\maketitle

\begin{abstract}
Text classification, an integral task in natural language processing, involves the automatic categorization of text into predefined classes. Creating supervised labeled datasets for low-resource languages poses a considerable challenge. Unlocking the language potential of low-resource languages requires robust datasets with supervised labels. However, such datasets are scarce, and the label space is often limited.

In our pursuit to address this gap, we aim to optimize existing labels/datasets in different languages. This research proposes a novel perspective on Universal Cross-Lingual Text Classification, leveraging a unified model across languages. Our approach involves blending supervised data from different languages during training to create a universal model. The supervised data for a target classification task might come from different languages covering different labels. The primary goal is to enhance label and language coverage, aiming for a label set that represents a union of labels from various languages.
We propose the usage of a strong multilingual SBERT as our base model, making our novel training strategy feasible. This strategy contributes to the adaptability and effectiveness of the model in cross-lingual language transfer scenarios, where it can categorize text in languages not encountered during training. Thus, the paper delves into the intricacies of cross-lingual text classification, with a particular focus on its application for low-resource languages, exploring methodologies and implications for the development of a robust and adaptable universal cross-lingual model.
\end{abstract}


\section{Introduction}
In the realm of Natural Language Processing, the creation of robust labeled datasets is crucial for effective text classification \cite{joshi2020deephindi,kulkarni2022experimental}. However, low-resource languages face challenges due to a scarcity of annotated corpora, lexicons, and grammar resources, limiting the variety and depth of linguistic labels \cite{joshi2022l3cube_mahanlp}.

This constraint hampers model training, affecting their ability to capture nuanced language variations. Additionally, the available datasets often have a very limited target label space, further complicating the task. As a result, text classification performance suffers in low-resource languages, impacting downstream applications. To address these challenges and enhance models in such scenarios, it becomes essential to not only expand datasets but also improve label coverage.

\begin{figure}[h]
    \centering
    \includegraphics[width=\linewidth]{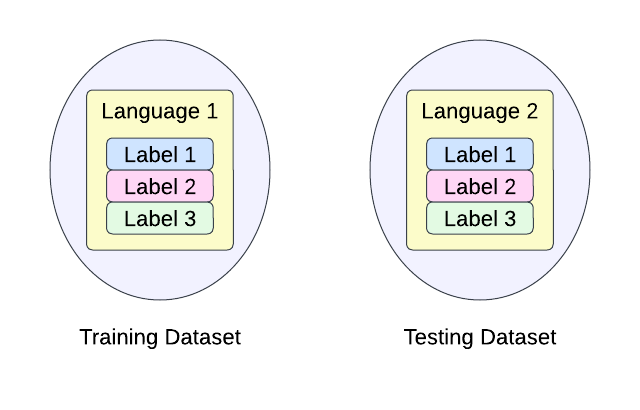}
    \caption{Cross Lingual Text Classification}
    \label{fig:traditional}
\end{figure}

\begin{figure}[h]
    \centering
    \includegraphics[width=\linewidth]{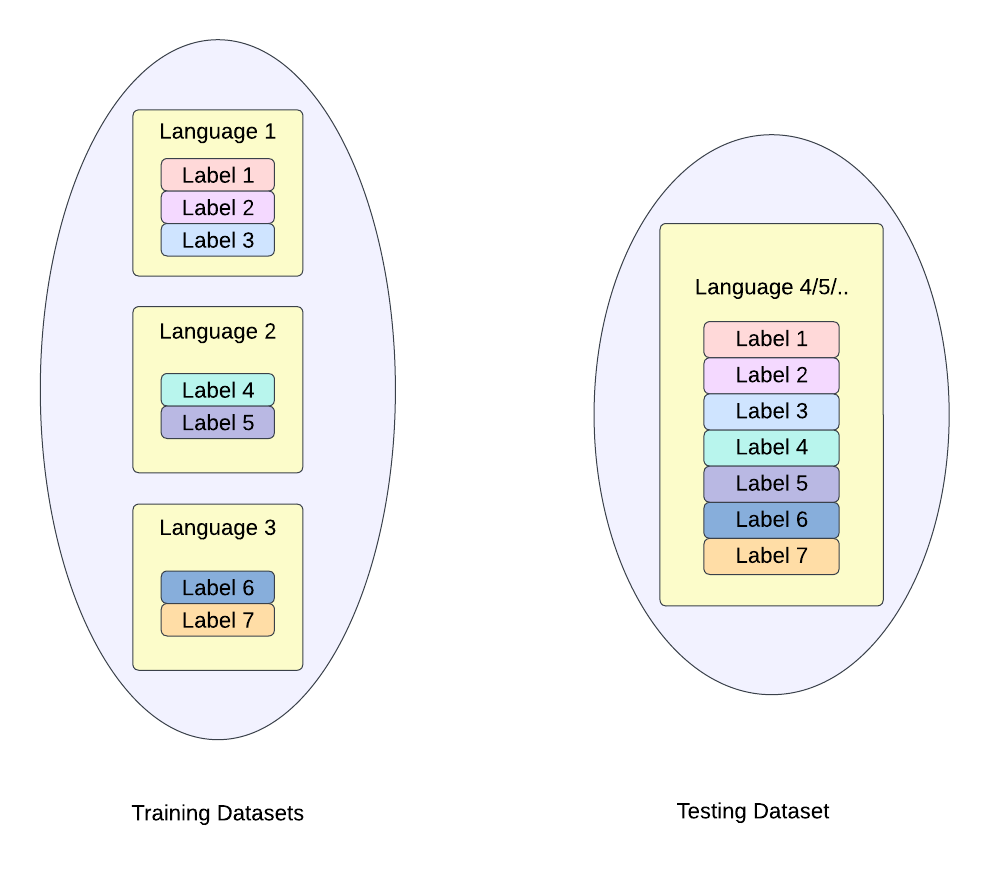}
    \caption{Representation of Universal Cross Lingual Text Classification}
    \label{fig:universal}
\end{figure}

Cross-lingual text classification provides a practical solution to address challenges stemming from limited data in low-resource languages. It enables the classification of text in unfamiliar languages not encountered during training. By utilizing models trained on more abundant languages, cross-lingual methodologies transfer knowledge to low-resource languages \cite{karamanolakis2020cross,dong-de-melo-2019-robust}. Various multilingual models \cite{devlin2019bert,feng2022languageagnostic,deode2023l3cubeindicsbert}, demonstrate robust performance in cross-lingual scenarios, effectively conducting text classification in low-resource, unseen languages after training on high-resource languages (refer to Fig \ref{fig:traditional}). However, even with this approach, the target labels are still confined to the single language data used for training. Each language typically possesses its unique set of labels, as illustrated in Table \ref{tab:dataset_details}, creating a challenge in developing a universally applicable model.

We present a novel universal approach for cross-lingual text classification. A universal model, in our context, is one that supports all languages and labels. Our method addresses the challenges posed by limited labeling in low-resource languages by combining datasets with diverse languages, each supporting a variety of labels. This amalgamation aims to train a universal model capable of cross-lingual text classification with a broader label coverage, fostering inclusivity and adaptability. The strategy is visually depicted in Fig \ref{fig:universal}, illustrating the incorporation of diverse labels from various languages during model training, enabling seamless operation across a range of languages and offering a more adaptable solution.

Our approach hinges on utilizing a robust multilingual Sentence-BERT (SBERT) as the foundation of our model. We evaluate the model's capabilities and enhance them through our proposed Universal Cross-Lingual Text Classification strategy. The trained universal model works well for all the languages that the original multilingual SBERT model supports. The classification model also supports all the labels used during training, regardless of the language of the corresponding sample text. In our pursuit of developing a comprehensive cross-lingual universal model, experiments were conducted using various BERT and SBERT models. Initially, we assess the capability of multilingual SBERT models for cross-lingual text classification, demonstrating their proficiency on unseen languages like Marathi despite being trained on other Indic languages. This observation highlights the model’s effectiveness in cross-lingual text classification within a specific language family from the Indian subcontinent. Subsequently, we evaluate our proposed approach by incorporating labeled data from different languages during training, aiming to create a universal model for cross-lingual text classification with wider label coverage.  

\section{RELATED WORK}
%introduction of related works
The relevant work in the field of cross-lingual text classification encompasses studies focusing on low-resource languages, limited corpora/datasets, various methodologies employed for cross-lingual text classification, and the evolution of models that have demonstrated improvements in this context. Researchers have explored solutions tailored to languages with scarce linguistic resources, devised methods to handle limited datasets effectively, and advanced models that enhance cross-lingual text classification performance.

%Low Resource
Addressing language modeling challenges for low-resource languages, where textual data is limited \cite{adams-etal-2017-cross}, and phonemic transcriptions are often scarce, remains a critical concern. While neural network language models (NNLMs) are praised for their ability to handle sparse data, it remains uncertain if this advantage extends to extremely data-limited scenarios. Developing technologies for Low-Resource Languages (LRLs) is imperative, not only for economic growth but also to preserve languages and prevent extinction \cite{magueresse2020lowresource}.

%Cross Lingual
Methods for cross-lingual text classification often rely on expensive resources like large parallel corpora or machine translation systems \cite{mihalcea-etal-2007-learning} limiting their applicability to high-resource languages. To overcome this CLTS \cite{karamanolakis2020cross} employs a translation budget to select seed words while training in a high-resource language (LS). Subsequently, it transports both these words and the trained model to a less-resourced language (LT), establishing a teacher classifier that incorporates these translated words. A study \cite{dong-de-melo-2019-robust} introduces a self-learning
framework that begins by learning from English samples and then expands its knowledge through predictions on unlabeled non-English samples.

Similarly, \cite{conneau2020unsupervised} introduces a comprehensive study on large-scale unsupervised cross-lingual representation learning. The authors propose a novel approach, utilizing massive amounts of multilingual data and introducing a translation language modeling objective, to train models capable of learning language-agnostic representations. The paper demonstrates the effectiveness of the proposed method across various cross-lingual benchmarks and tasks, highlighting its scalability and success in capturing nuanced linguistic relationships.

In the landscape of cross-lingual representation learning, LaBSE (Language-agnostic BERT Sentence Embedding) and LASER (Language-Agnostic SEntence Representations) stand out as noteworthy contributions. LaBSE \cite{feng2022languageagnostic} focuses on language-agnostic sentence embedding by pre-training BERT on parallel data from multiple languages, demonstrating its efficacy in cross-lingual applications such as sentence retrieval and document classification. Similarly, LASER \cite{Artetxe_2019}, offers a language-agnostic approach by training on parallel data in 93 languages. LASER's embedding have shown utility in tasks like cross-lingual information retrieval and sentiment analysis. Both LaBSE and LASER address the challenge of handling diverse languages, providing valuable insights and techniques for cross-lingual representation learning. But, the drawbacks in LASER encompass a large model size, potential disparities in data quality, reliance on pre-trained models, and a limitation in capturing fine-grained sentence details. 

% The selection between LASER and Transformer-based models, such as XLM-RoBERTa \cite{conneau2020unsupervised} , hinges on the specific requirements of the task and the characteristics of the utilized data.

BERT \cite{devlin2019bert} is a bidirectional transformer model that learns contextual language representations through masked language model pre-training. Sentence-BERT improves upon BERT by specifically focusing on encoding entire sentences rather than individual words, yielding better sentence-level embeddings for various natural language processing tasks. This targeted approach enhances performance in tasks requiring sentence-level understanding. To leverage the benefits of Sentence-BERT (SBERT), our study employs the L3Cube-IndicSBERT model \cite{deode2023l3cubeindicsbert}. This SBERT model harnesses the capabilities of multilingual BERT to facilitate efficient learning of sentence representations that capture semantic meaning across diverse languages. This work contributes to the landscape of cross-lingual representation learning by offering a straightforward yet effective approach, demonstrating the accessibility and utility of multilingual BERT for cross-lingual tasks.

\section{Methodologies}\label{sec:Methodologies}

Various multilingual models have demonstrated robust cross-lingual performance. Trained on distinct labels of one language, these models accurately predict labels of a different language, enabling cross-lingual text classification on an unseen language Fig \ref{fig:traditional}. This motivates our approach for Universal Cross-Lingual Text Classification, which involves training a model with supervised labels from different languages and testing it on a completely unfamiliar language Fig \ref{fig:universal}.

\begin{figure}[h]
    \centering
    \includegraphics[width=\linewidth]{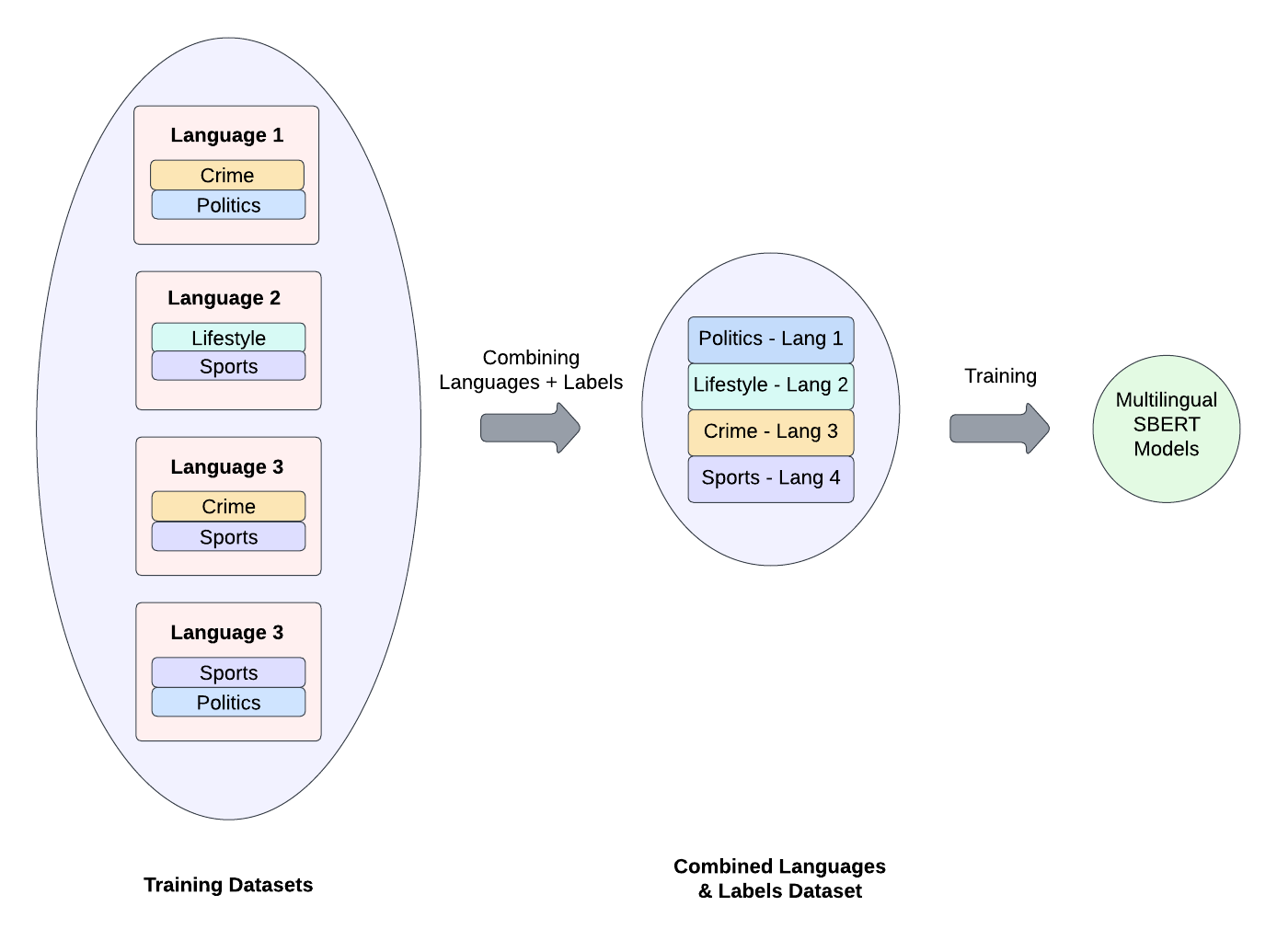}
    \caption{Proposed approach for Universal Cross Lingual Text Classification}
    \label{fig:methodology}
\end{figure}

To enhance the prediction of labels for unfamiliar languages, we employ a technique that merges unique labels from various languages into a unified dataset, subsequently using this combined dataset for model training [Fig \ref{fig:methodology}]. The merging of labels and languages can be customized depending on the availability of label sets. As illustrated in the figure, we gathered all four distinctive labels from diverse languages. However, the creation of a universal model extends beyond this specific approach. We can simply pool all the training data together without considering language or target labels. In this work, as an extreme case, we consider unique labels across languages during training and unseen languages during evaluation. However in general we need not restrict to unseen languages or unique labels.
%It allows for the combination of a single label, such as 'Sports,' from all languages, contributing to a more effective and inclusive model training process.

Selecting an appropriate model is crucial hence we compare the cross-lingual performances of different models and opt for the one with the highest accuracy scores. Our experimental procedure involves obtaining sentence embeddings from an SBERT model and then using a classifier on top of it for classification. To show the high performance of the model, we use the trained classifier to predict some labels in an entirely unfamiliar language, not used during the training of the classifier. This step assesses the model's adaptability and accuracy in unfamiliar linguistic contexts, enhancing overall flexibility.

Specifically, we use sentence encoder (SBERT) models to extract sentence embeddings, followed by the K-Nearest Neighbors Algorithm for the classification of these embeddings into predefined classes. Our experiment systematically explores the 'k' parameter (1 to 100), determining the optimal 'k' value by evaluating the highest accuracy on the validation dataset. For our experiment, we use the IndicNLP News Article Classification dataset, comprising news articles from nine distinct Indic languages categorized into different classes [Table \ref{tab:dataset_details}]. Our classification models include both BERT- based models and LASER.

\subsection{MODELS \& DATASET}
This section provides a detailed exposition of the diverse models and dataset employed in our experiment.
\subsubsection{IndicSBERT\textsuperscript{\cite{deode2023l3cubeindicsbert}}}
IndicSBERT \footnote{\url{https://huggingface.co/l3cube-pune/indic-sentence-bert-nli}},is a multilingual SBERT model for 10 Indian regional languages .The IndicSBERT uses Multilingual Representations for Indian Languages (MuRIL) as the base model. IndicSBERT model  \footnote{\href{https://huggingface.co/l3cube-pune/indic-sentence-bert-nli}{l3cube-pune/indic-sentence-bert-nli}} trained on the Semantic Textual Similarity (STS) dataset encompassing ten major Indian languages, this singular model demonstrates proficiency across English,  Kannada, Marathi, Hindi, Tamil,  Gujarati, Telugu,  Punjabi, Oriya, Bengali and Malayalam. Furthermore, the model exhibits cross-lingual capabilities, allowing it to effectively process and understand diverse languages beyond its training set. It performs really well on Indic cross-lingual and monolingual sentence similarity tasks. 

\subsubsection{LaBSE\textsuperscript{\cite{feng2022languageagnostic}}} 
LaBSE\footnote{\url{https://huggingface.co/sentence-transformers/LaBSE}} a multilingual BERT embedding model \cite{feng2022languageagnostic}, can produce language-agnostic cross-lingual sentence embedding for 109 languages. It excels in identifying sentence translations and is referred to as Language-agnostic BERT sentence embedding model that combines the best methods from Masked Language Modeling (MLM) and Translation Language Modeling (TLM) to improve the performance of sentence embedding. This multilingual BERT embedding model demonstrates its proficiency during training by generating vectors that closely align for bilingual sentence pairs translating each other. By emphasizing the proximity of vectors in expressing the semantic equivalence of translated phrases, LaBSE showcases its effectiveness. Thus, LaBSE can be used for a variety of tasks, such as cross-lingual natural language inference, cross-lingual sentiment analysis, and cross-lingual machine translation.

\subsubsection{LASER\textsuperscript{\cite{Artetxe_2019}}}
Facebook introduced LASER\footnote{\url{https://github.com/facebookresearch/LASER}} (Language-Agnostic Sentence Representations), which provides multilingual sentence representations for 93 languages, including Marathi and Hindi, which have limited linguistic resources. It generates language-agnostic sentence embeddings across 93 languages, supporting cosine similarity for sentence comparison. It is noteworthy as it utilises a single model that can handle several languages and cross-lingual transferability. By embedding all languages collectively into a common space, this paradigm achieves language integration and allows for a comprehensive yet language-neutral approach. 
%The drawbacks for LASER encompass a large model size, potential disparities in data quality, reliance on pre-trained models, and a limitation in capturing fine-grained sentence details. The selection between LASER and Transformer-based models, such as XLM-RoBERTa \cite{conneau2020unsupervised}, hinges on the specific requirements of the task and the characteristics of the utilized data.

\subsubsection*{DATASET: Indic-NLP News Articles \footnote{\url{https://github.com/AI4Bharat/indicnlp_corpus}}}
It is a set of classification datasets, encompassing news articles and their corresponding categories across nine Indian languages. The dataset maintains balance among labels for news articles. Table \ref{tab:dataset_details}  presents the exact number of samples for training, testing and validation for the dataset, along with the labels for every language.\\

\begin{table}[h]
    \centering
    \resizebox{\columnwidth}{!}{
    \begin{tabular}{lcccc}
        \toprule
        \textbf{Language} & \textbf{Train} & \textbf{Test} & \textbf{Valid} & \textbf{Labels} \\
        \midrule
        Bengali (bn) & 11200 & 1400 & 1400 & entertainment, sports \\
        Gujarati (gu) & 1632 & 204 & 204 & business, entertainment, sports \\
        Kannada (kn) & 24000 & 3000 & 3000 & entertainment, lifestyle, sports \\
        Malayalam (ml) & 4800 & 600 & 600 & business, entertainment, sports, technology \\
        Marathi (mr) & 3815 & 478 & 477 & entertainment, lifestyle, sports \\
        Oriya (or) & 24000 & 3000 & 3000 & business, crime, entertainment, sports \\
        Punjabi (pa) & 2496 & 312 & 312 & business, entertainment, sports, politics \\
        Tamil (ta) & 9360 & 1170 & 1170 & entertainment, politics, sport \\
        Telugu (te) & 19200 & 2400 & 2400 & entertainment, business, sports \\
        \bottomrule
        
    \end{tabular}}
    \caption{Training, testing, validation and labels for 9 Indian Languages}
    \label{tab:dataset_details}
\end{table}

\subsection{EXPERIMENT}
We evaluated multilingual models for their effectiveness in cross-lingual text classification. The initial focus was on LASER, LaBSE, and IndicSBERT, evaluating their performance using cross-lingual text classification on the diverse IndicNLP dataset spanning nine languages. The second experiment explored model generalization and linguistic adaptability by training on a subset of nine languages with varied label combinations and testing on four distinct languages excluded from the training set. This approach allowed us to evaluate the model's cross-lingual performance and its proficiency in label prediction for languages unseen during training, addressing diverse linguistic challenges.

\subsubsection{Cross-lingual text classification}{\label{Experiment 1}}

To assess the effectiveness of multilingual models—specifically LASER, LaBSE, and IndicSBERT, IndicSBERT-STS we conducted a comprehensive analysis using cross-lingual text classification. In the evaluation process, we measured the accuracy of each model.

The categories within the IndicNLP dataset differ across all nine languages, with the only common labels being 'entertainment' and 'sports.' To maintain sample uniformity, we selected an equal number of samples and ensured shuffling to prevent repeated occurrences of the same labels. Consequently, we conducted cross-lingual text classification by individually training 1000 samples for Bengali, Kannada, Marathi, Tamil, Oriya, Malayalam and Telugu. The testing phase on Marathi involved combining the 'entertainment' and 'sports' labels from the validation and test datasets. In consideration of the limited number of training samples for Gujarati and Punjabi, we trained 500 samples each. Thus, all four models were trained on identical samples to assess precise accuracy without variations in the dataset.

\begin{figure}[h]
    \centering
    \includegraphics[scale=0.45]{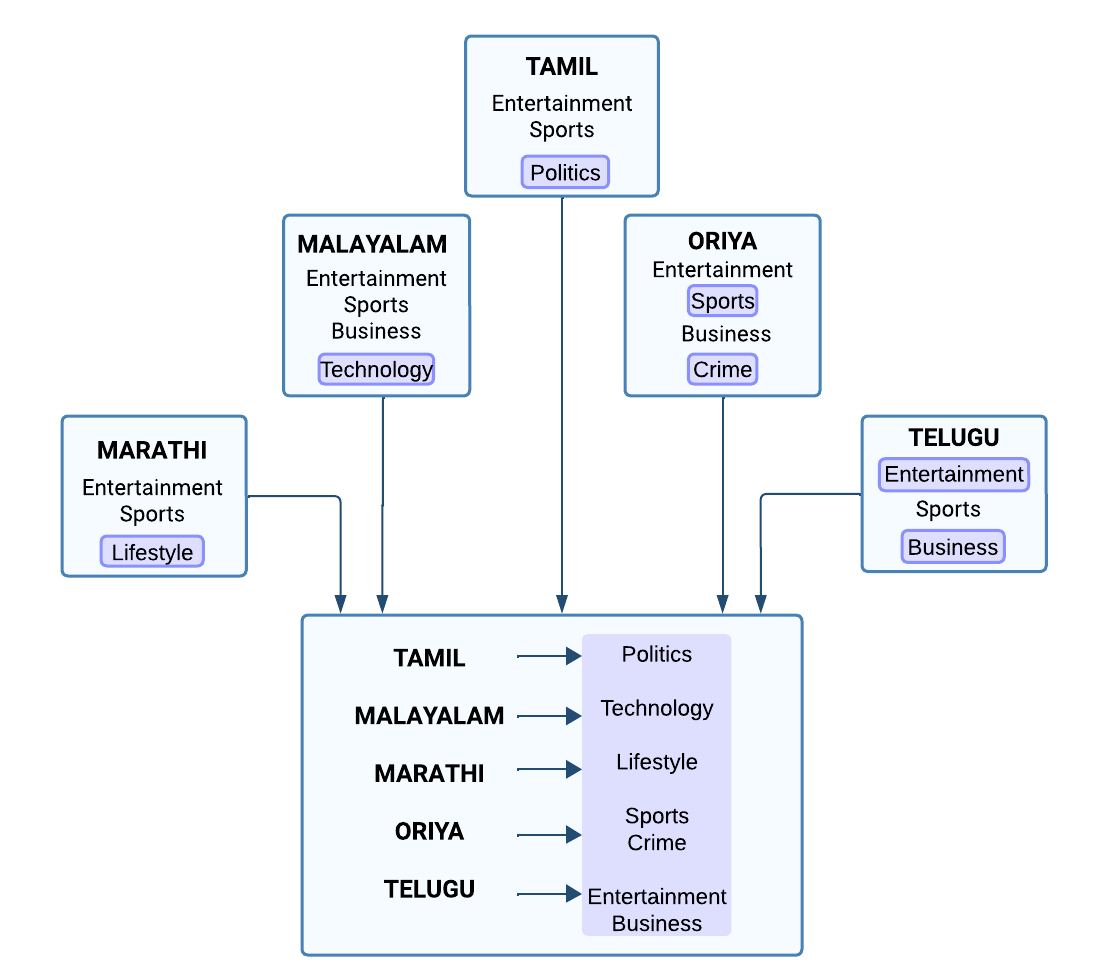}
    \caption{Combining of labels across 5 languages for model training}
    \label{fig:Mixture of Dataset}
\end{figure}

\subsubsection{Universal cross-lingual text classification}\label{Experiment 2}

As indicated in Table \ref{tab:dataset_details}, there is insufficient label coverage across all languages, with some languages possessing certain labels while others lack them. To address this, the experimental setup involves training on a subset of 9 languages, utilizing 5 for training with various label combinations, while the remaining 4 are designated for testing purposes. This approach aims to assess the model's ability to generalize and accurately perform text classification across a diverse set of languages with different labels. Thereby it provides a comprehensive evaluation of its linguistic adaptability, and meets the primary objective of cross-lingual text classification, signifying the capability to accurately classify text in languages unseen during the model's training phase.

\begin{table*}[b]
   \centering
    \begin{tabular}{|l|*{9}{c|}}
    \hline
        \textbf{Training Language/Model} & \textbf{Bengali} & \textbf{Kannada} & \textbf{Marathi} & \textbf{Tamil} & \textbf{Telugu} & \textbf{Malayalam} & \textbf{Oriya} & \textbf{Punjabi} & \textbf{Gujarati} \\
        \hline
            IndicSBERT-STS & 0.9792 & 0.9792 & 0.9762 & 0.9911 & 0.9702 & 0.9643 & 0.9851 & 0.9732 & 0.9673 \\
            IndicSBERT & 0.9732 & 0.9881 & 0.9821 & 0.9792 & 0.9762 & 0.9762 & 0.9732 & 0.9821 & 0.9732 \\
            LaBSE & 0.9673 & 0.9851 & 0.9792 & 0.9792 & 0.9762 & 0.9732 & 0.9792 & 0.9762 & 0.9792 \\
            LASER & 0.8512 & 0.5536 & 0.9435 & 0.7649 & 0.8244 & 0.8065 & 0.4613 & 0.5298 & 0.4673 \\
        \hline
    \end{tabular}
\caption{Accuracy scores for 'entertainment' and 'sports' labels across Indic languages, with a focus on Marathi testing}
\label{tab:Table 1} 
\end{table*}

For the experimental setup, the label combinations for the training model are derived from the Indic-NLP news articles dataset. The mixture of labels ensures that the model is exposed to a wider and comprehensive array of linguistic variations, enhancing its capacity for cross-lingual text classification. As represented in Fig. \ref{fig:Mixture of Dataset} the entertainment label is sourced from Telugu articles, sports from Oriya, business from Telugu, lifestyle from Marathi, technology from Malayalam, crime from Oriya, and politics from Tamil. In order to ensure homogeneity in the samples, we have chosen 1000 samples from each label and implemented shuffling to prevent consecutive occurrences of the same labels. This selection aims to create a diverse and representative training set, ensuring exposure to various linguistic and thematic nuances present in different Indian languages. The experiment is performed on two selected models, IndicSBERT and LaBSE for further comparison.

In the testing phase, we conduct performance evaluation of complete datasets of four distinct languages : Bengali, Kannada, Gujarati, and Punjabi that were intentionally excluded from the model's training dataset. This deliberate selection of testing languages serves to assess the model's generalization ability and its aptitude for cross-lingual text classification across languages that were not part of its initial training regimen. The chosen testing languages present unique linguistic challenges, allowing for a robust evaluation of the model's cross-lingual performance.

\section{RESULTS}
In Experiment 1 (Section \ref{Experiment 1}) we conducted a cross-lingual text classification task utilizing the IndicNLP dataset across various Indic languages, focusing on the specific labels 'entertainment' and 'sports'. Subsequently, Table \ref{tab:Table 1} presents the accuracy scores obtained when the model was tested on Marathi. The accuracy scores are notably high, attributed to the substantial relatedness among the Indic languages. LaBSE, IndicSBERT-STS and IndicSBERT outperform the LASER model. Following that, in Fig. \ref{fig:Datasets}, a visual comparison of accuracy scores is presented across various languages and models, based on testing with the Marathi dataset. In Fig. \ref{fig:Datasets}, the horizontal axis illustrates the training languages. The accuracy surpass 0.97, even when the model undergoes testing on a language it has not been trained on previously. The IndicSBERT model consistently yields superior accuracy when compared to other models. Consequently, we have chosen to employ IndicSBERT for our subsequent tasks. This decision is driven by the model's demonstrated proficiency in delivering high accuracy, making it a reliable choice for further experiments within our research framework.

\begin{figure}[h]
    \centering
    \includegraphics[width=\linewidth]{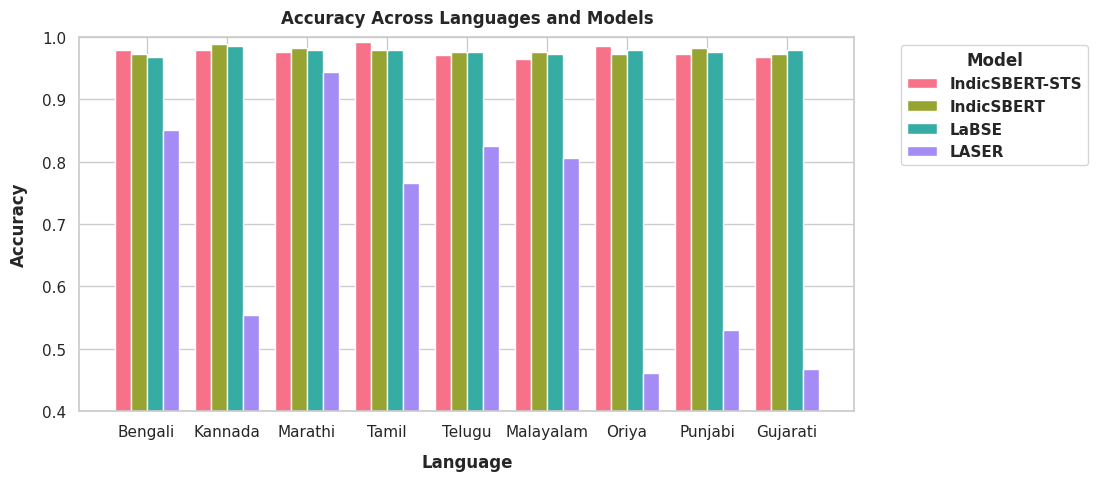}
    \caption{Comparison of accuracy scores obtained from various models on Marathi test set. \newline 
    Horizontal axis depicts the training languages.}
    \label{fig:Datasets}
\end{figure}

In Experiment 2 (Section \ref{Experiment 2}), we performed testing on four distinct languages that were excluded from the training set. Utilizing the IndicSBERT and LaBSE models, the evaluation employed the K-Nearest Neighbors (KNN) algorithm, a machine learning approach for classification. The outcomes of the KNN-based assessments revealed remarkably high accuracy levels.

The outcomes presented in Table \ref{tab:performance_comparison} depict the performance evaluation conducted by testing the complete test datasets of Bengali, Kannada, Gujarati, and Punjabi separately on the trained model. Additionally, the combined testing involved merging the datasets of all four languages and assessing the model's performance on this unified dataset. Although this combined testing dataset is substantial, it aligns with the overarching research focus on cross lingual text classification.  Based on the results obtained, it is evident that IndicSBERT outperforms LaBSE in many scenarios. Thus IndicSBERT exhibits strong multilingual and cross-lingual capabilities in most of the observations and is one of the prominent models to perform universal cross lingual text classification.

The obtained results underscore the importance of enhancing the model's proficiency in languages with limited linguistic resources. This emphasis addresses the core objective of achieving cross-lingual text classification, particularly in scenarios where languages have sparse available data.

\begin{table}[h] 
  \begin{center}
    \begin{tabular}{|c|c|c|}
      \hline
      \textbf{Testing Language} & \textbf{IndicSBERT(Accuracy)} & \textbf{LaBSE(Accuracy)} \\
      \hline
      Bengali & 0.9336 & 0.8979 \\
      Gujarati & 0.9412 & 0.8529 \\
      Kannada & 0.8760 & 0.9310 \\
      Punjabi & 0.9071 & 0.9359 \\
      \hline
      \textbf{All Combined} & \textbf{0.9035} & \textbf{0.8999} \\
      \hline

    \end{tabular}
    
  \end{center}
      \caption{Accuracy score obtained from IndicSBERT and LaBSE after testing on 4 Languages individually and combined}
    \label{tab:performance_comparison}
\end{table}

\section{Conclusion}\label{sec:Conclusion}
\noindent
Our research tackles the challenges posed by limited datasets and the non-uniform distribution of supervised data across diverse Indic languages. Through a novel approach of blending supervised data from different languages and subsequently testing on previously unseen languages, we demonstrate a significant increase in label coverage. This  methodology underscores the potential for testing on languages irrespective of whether the same language was used in training, presenting a valuable contribution to the field of cross-lingual text classification. Our research specifically evaluates the IndicSBERT model, comparing its performance with widely used models such as LASER and LaBSE across various training scenarios. The outcomes demonstrate the superiority of IndicSBERT in the realm of universal cross-lingual text transfer. Our study tackles the issue of limited labeled data in low-resource languages and underscores the versatility of the IndicSBERT model. By doing so, our research provides valuable perspectives and pragmatic solutions for achieving universal cross-lingual text classification in the context of low-resource languages.\\

\section*{Acknowledgements} 
\noindent This work was completed as part of the L3Cube Mentorship Program in Pune. We would like to convey our thankfulness to our L3Cube mentors for their ongoing support and inspiration.

\printbibliography
\end{document}